\documentclass[runningheads]{llncs}
\usepackage{graphicx}

\usepackage{xcolor}
\usepackage{amsmath}
\usepackage{dirtytalk}
\usepackage{nicefrac}

\begin{document}
\title{Towards Robust General Medical Image Segmentation}

\author{Laura Daza$^1$ \and Juan C. P\'erez$^{1,2}$ \and Pablo Arbel\'aez$^1$}

\authorrunning{L. Daza et al.}

\institute{$^1$Universidad de los Andes, Colombia\\
$^2$King Abdullah University of Science and Technology (KAUST), Saudi Arabia}

\maketitle              
\begin{abstract}
The reliability of Deep Learning systems depends on their accuracy but also on their robustness against adversarial perturbations to the input data. Several attacks and defenses have been proposed to improve the performance of Deep Neural Networks under the presence of adversarial noise in the natural image domain. However, robustness in computer-aided diagnosis for volumetric data has only been explored for specific tasks and with limited attacks. We propose a new framework to assess the robustness of general medical image segmentation systems. Our contributions are two-fold: \textit{(i)} we propose a new benchmark to evaluate robustness in the context of the Medical Segmentation Decathlon (MSD) by extending the recent \textit{AutoAttack} natural image classification framework to the domain of volumetric data segmentation, and \textit{(ii)} we present a novel lattice architecture for RObust Generic medical image segmentation (ROG). Our results show that ROG is capable of generalizing across different tasks of the MSD and largely surpasses the state-of-the-art under sophisticated adversarial attacks.

\keywords{Robustness assessment \and adversarial training \and adversarial attacks \and general medical segmentation}
\end{abstract}
\section{Introduction}

The observation that imperceptible changes in the input can mislead Deep Neural Networks (DNN)~\cite{szegedy2014intriguing,carlini2017towards} has attracted great interest in the deep learning community. This behaviour has been studied for various tasks, such as classification~\cite{szegedy2014intriguing,madry2018towards}, object detection~\cite{zhang2019detection} and semantic segmentation~\cite{xie2017adversarial,arnab2018robustness,cisse2017houdini}, highlighting the importance of \textit{reliably} assessing robustness. To improve this dimension, adversarial robustness has been studied both from the side of the attacks~\cite{moosavi2016deepfool,brendel2018decision,croce2020reliable,ozbulak2019impact,Joeladvattack2021} and the defenses~\cite{zhang2019theoretically,shafahi2019adversarial,mummadi2019defending,liu2020defending}, obtaining sizable progress towards DNN models resistant to adversarial perturbations; however, still much is left to advance. In particular, recognition tasks in medical domains are of utmost importance for robustness, as these tasks aim at safety-critical applications in which brittleness could have dramatic consequences.

Most semantic segmentation methods for computer aided-diagnosis are designed for specific anatomical structures~\cite{Vnet,zhu20183d,isensee2020nnu,tang2019nodulenet}. This level of specialization has been attributed to the large variability in structures, acquisition protocols, and image modalities, the limited annotated datasets, and the computational cost of processing 3D data. The Medical Segmentation Decathlon (MSD)~\cite{decathlon}, an experimental framework that combines ten diagnostic tasks, aims at addressing this over-specialization. Top-performing techniques in the MSD solve each task by combining multiple models and automatically adjusting task-specific parameters~\cite{nnunet}, or conducting Neural Architecture Search (NAS)~\cite{zhu2019v,c2fnas}. The MSD, thus, is a promising benchmark to develop general methods for medical image segmentation. However, adversarial robustness, a critical dimension of deep learning systems, remains uncharted territory in the MSD.

\begin{figure}[t]
\begin{center}
\includegraphics[width=0.71\linewidth, trim={0 0 0 1.4cm}, clip]{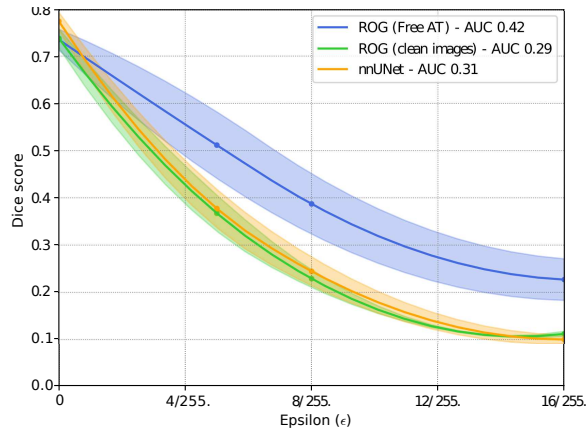}
\end{center}
\caption{\textbf{Medical Segmentation Robustness Benchmark.} We introduce a new benchmark for studying adversarial robustness on the MSD~\cite{decathlon}. We propose a novel method for robust general segmentation that significantly outperforms the state-of-the-art in our benchmark~\cite{nnunet}, providing a strong baseline for future reference.}
\label{fig:main}
\end{figure}

In this paper, we propose a new experimental framework to evaluate the adversarial robustness of medical segmentation methods. With this aim, we build on the AutoAttack benchmark~\cite{croce2020reliable}, an ensemble of adversarial attacks to evaluate recognition models in natural images. We extend AutoAttack from the image classification domain to volumetric multi-channel semantic segmentation. Our results suggest that the adversarial vulnerability of methods in the medical domain, while previously evidenced~\cite{ma2020understanding,li2019volumetric}, has most likely been \textit{underestimated}. We find that subjecting segmentation methods to our extension of AutoAttack exposes their brittleness against adversarial perturbations. 

As a strong baseline for RObust General (ROG) medical segmentation, we propose an efficient lattice architecture that segments organs and lesions on MRI and CT scans. Further, we leverage Adversarial Training (AT)~\cite{goodfellow2015explaining,madry2018towards} to protect our models against adversarial perturbations. Since AT's great computational demands can render training of 3D models prohibitive, we equip ROG with \say{Free} AT~\cite{shafahi2019adversarial}, an efficient version of AT. Figure \ref{fig:main} shows the variation in Dice score as the attack's strength, $\epsilon$, increases, comparing ROG trained with clean images, trained with Free AT and nnUnet~\cite{nnunet}, the winner of the MSD challenge.

Our main contributions can be summarized as follows: \textit{(i)} we extend AutoAttack to the domain of 3D segmentation to introduce an experimental framework for evaluating adversarial robustness in medical segmentation tasks, and \textit{(ii)} we introduce ROG, the first generic medical segmentation model that is robust against adversarial perturbations. We will make our code publicly available to ensure reproducibility and to encourage robustness assessments as a fundamental dimension in the evaluation of 3D DNNs in safety-critical applications\footnote{https://github.com/BCV-Uniandes/ROG}.

\section{Methodology}

\subsection{Adversarial robustness}
\subsubsection{Extending \textit{AutoAttack} to 3D segmentation.}
Assessing adversarial robustness has proven to be a complex task~\cite{carlini2019evaluating}, hence, many approaches have been proposed~\cite{carlini2017towards,moosavi2016deepfool}. In this work, we build on AutoAttack~\cite{croce2020reliable}, a parameter-free ensemble of attacks that has shown remarkable capacity for breaking adversarial defenses and has emerged as the leading benchmark to assess adversarial robustness. AutoAttack is composed of: \textit{(i)} AutoPGD-CE and \textit{(ii)} AutoPGD-DLR, two Projected Gradient Descent-based (PGD) adversaries that optimize Cross Entropy and Difference-of-Logits Ratio~\cite{croce2020reliable} respectively, \textit{(iii)} targeted Fast Adaptive Boundary (FAB-T)~\cite{croce2020minimally} which minimizes a decision-changing perturbation, and \textit{(iv)} Square Attack~\cite{andriushchenko2020square}, a score-based black-box attack.

Image classification is the most studied setting for adversarial robustness~\cite{moosavi2017universal,madry2018towards}. Thus, AutoAttack was designed to fool models that assign a probability distribution to \textit{each image}. However, we are interested in attacking models that assign a probability distribution to \textit{each spatial location}. Therefore, we generalize the definition of the attacks to consider both input and output as four-dimensional tensors: three spatial dimensions plus the channels and the spatial dimensions plus the probability distribution over the classes. In addition, every attack requires a notion of \say{success}, \textit{i.e.} the state in which the attack has fooled the model. While the success of an attack is readily defined for classification, it is not clearly defined for segmentation. Thus, we re-define this notion as a drop in the Dice score below $\nicefrac{\mu_{i}}{2}$, where $\mu_{i}$ is the average performance in task $i$ over clean images. With this definition, we modify each attack to compute voxel-wise functions across all spatial locations. Finally, we preserve the definition of \say{Robust Accuracy} as the per-instance worst case across the four attacks.

\subsubsection{Adversarial training}.
Given the critical applications of medical image segmentation, it is crucial to develop methods that are both accurate and robust to perturbations. To improve ROG's robustness, we apply \say{Free} Adversarial Training~\cite{shafahi2019adversarial} in our task domain to fine-tune our models, following the spirit of~\cite{gupta2020improving,cai2018curriculum}. For Free AT, we set $\epsilon$, the strength of the attack, to $\nicefrac{8}{255}$, and $m$, the number of times an image is replayed for gradient computation, to $5$.

\begin{figure}[t]
\begin{center}
\includegraphics[width=0.7\linewidth]{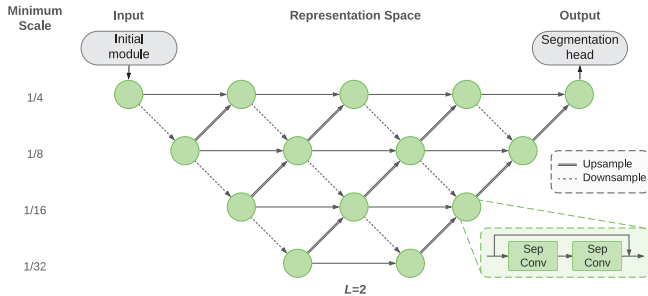}
\end{center}
   \caption{\textbf{Overview of ROG}. Our lattice architecture for generic medical segmentation preserves high resolution features while also exploiting multiple image scales. The initial module adjusts the input by reducing its scale to a minimum of \nicefrac{1}{4} of the original size, while the segmentation head adapts to produce a mask of the same size as the input.}
\label{fig:gnet}
\end{figure}

\subsection{Generic Medical Segmentation}
We aim to study the unexplored dimension of adversarial robustness in general medical image segmentation. However, existing methods lack publicly available resources~\cite{c2fnas} or require training multiple models for each task~\cite{nnunet}. Combined with the high computational demands of adversarial training, these limitations prevent their adoption for our primary goal.
Hence, we propose ROG, an efficient single-architecture model for \textit{RObust Generic} segmentation that exploits increasingly larger receptive fields while preserving high-resolution features. ROG comprises an initial module with four convolutions, the main lattice of processing nodes, and a segmentation head with two convolutions. We organize the nodes in a triangular lattice with four scales, as shown in Figure~\ref{fig:gnet}. Unlike UNet++~\cite{Unet++}, we connect each node with both upper and lower level nodes, and we remove the dense connections. As a result, ROG has 10$\times$ fewer parameters ($2.6M$ vs. $26.2M$) than~\cite{Unet++}. We find that our topology is significantly more efficient than a squared lattice since it reduces the number of layers while preserving the advantages of simultaneous multi-scale processing. We also observe that varying the length $L$ of the lattice, \textit{i.e.} the number of nodes in the last scale, benefits different types of objects. In particular, lower lengths tend to benefit small objects but are detrimental for large objects, while longer lengths favor the opposite. We set $L=2$ as a trade-off between accuracy and model size.

Within each node, we use two separable convolutions, instance normalization layers, and residual connections. We also use swish activations~\cite{swish}, given the benefit of its smooth gradient for adversarial training~\cite{swishYuille}. To connect nodes at different levels, we use $1\times 1\times 1$ convolutions to adjust the number of feature maps and change the spatial resolution by either using a stride of $2$ or adding a trilinear interpolation. Finally, we automatically adjust ROG for each task using the average size of the volumes in the dataset to select the input patch size, strides in the initial module, and upsampling factors in the segmentation head. This selection is driven by two main factors: the target objects' spatial dimensions and the model's memory consumption. Also, we make sure that the minimum possible spatial resolution is $\nicefrac{1}{32}$ of the input resolution.

\section{Experiments}

\subsection{Adversarial robustness assessment}
We train two variants of ROG for each task: using only clean images and using Free AT. We then use our extended version of AutoAttack to attack the models. For the AutoPGD and FAB attacks, we fix the number of iterations to $5$ and vary $\epsilon$ between $\nicefrac{5}{255}$ and $\nicefrac{16}{255}$ to assess the adversarial robustness of each model to perturbations of different magnitudes. However, since our experimental framework differs significantly from natural image classification, we also explore the attacks using between $5$ and $20$ iterations with $\epsilon=\nicefrac{8}{255}$ and report these results in the supplemental material. For the Square attack, we vary the number of queries between 500 and 5000. We train our models on 80\% of the data and validate on the remaining 20\%. To reduce bias during the evaluation, all design choices are made based on the results over the tasks of pancreas and prostate segmentation of the MSD. For reference, we evaluate the robustness of nnU-Net, the highest performing method on the MSD with publicly available resources.

\subsection{General medical segmentation}
We evaluate the generalization capacity of ROG on the MSD. Each task has a target that can be an organ or a tumor and source data that can be CT or MRI with one or multiple modalities. The number of patients varies between 30 and 750 for the tasks, with a total of 2,633 volumes~\cite{decathlon}. We use Adam~\cite{kingma2015adam} with a weight decay of $10^{-5}$, and an initial learning rate of $10^{-3}$ that is reduced by a factor of $0.5$ when the validation loss has stagnated for the previous $50$ epochs. We optimize a combination of the Dice loss and the Cross-Entropy loss. For data augmentation, we use random rotations, scaling, mirroring, and gamma correction. We sample cubic patches with a $50\%$ probability of being centered at a foreground category to address data imbalance. For pre-processing, we re-sample the volumes to ensure a homogeneous voxel spacing. Next, we follow~\cite{nnunet} and clip the intensities to the $[0.5,99.5]$ percentiles of the foreground values and perform $z$-score normalization.

For inference, we uniformly sample patches and combine the predictions to generate the volume. We combine the predictions of the same voxel in multiple patches via weights inversely proportional to the distance to the patch's center. To obtain the final predictions for the MSD test set, we perform simple noise reduction: for organs, we preserve only the largest element; for tumors, we fuse all elements smaller than a threshold with the surrounding category. The threshold is defined based on the average object size in the dataset.

\begin{figure}[!t]
\begin{center}
\includegraphics[width=0.7\linewidth, trim={0 0 0 1.4cm}, clip]{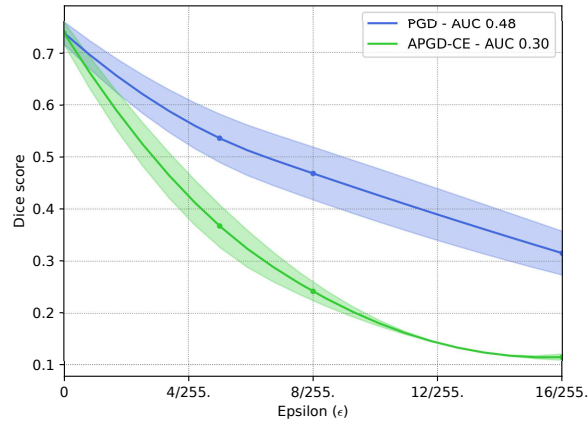}
\end{center}
   \caption{\textbf{PGD \textit{vs.} AutoPGD-CE.} We compare the performance of PGD and AutoPGD-CE by attacking ROG trained on clean images averaging the performance over the 10 tasks in the MSD. The results demonstrate that the AutoPGD-CE attack is significantly stronger than regular PGD.}
\label{fig:pgd}
\end{figure}

\section{Results}

\subsection{Adversarial robustness}
\subsubsection{Adversarial attacks.}
Previous assessments of adversarial robustness in segmentation tasks rely on the well-known PGD attack~\cite{mummadi2019defending}. To elucidate the need for stronger adversaries, we compare the performances of AutoPGD-CE and PGD. We run both attacks on our model trained with clean images in all the datasets, vary the attacks' strength ($\epsilon$), and report the average results in Figure~\ref{fig:pgd}. In line with~\cite{croce2020reliable}, we find that AutoPGD-CE is dramatically stronger for \textit{every} $\epsilon$. In particular, the AUC for PGD is $0.48$, while it drops to $0.30$ for AutoPGD-CE.

We test the individual attacks \textit{within} AutoAttack following the same configuration and report the average Dice scores in Table~\ref{tab:attacks}. We find that both PGD-based adversaries are the strongest attacks. Further, we note that FAB-T's performance is deficient: FAB-T is unable to find adversaries. While studies on FAB~\cite{croce2020minimally,croce2020reliable} often report lower performance for FAB when compared to PGD-based adversaries, our results present a stark contrast. While this observation calls for further experimentation, our initial hypothesis is that FAB's formulation, based on decision-boundary approximation through linearization near the input, may be ill-suited for segmentation. Thus, a re-definition of FAB's inner workings to account for the particularities of segmentation methods may be called for. Regarding the Square attack, despite the simplicity of the random search-based optimization, it consistently finds adversaries. However, since the search space for variable-sized cubes for volumetric inputs is rather large, the computational demands are notably higher than the other attacks.

\begin{table}[t]
\centering
\caption{\textbf{Comparison of individuals attacks of AutoAttack.} We report the average performance over all datasets in the MSD with the four attacks in AutoAttack.}
\begin{tabular}{c|c|c|c|c|c}
& \textbf{Clean} & \textbf{APGD-CE} & \textbf{APGD-DLR} & \textbf{FAB-T} & \textbf{Square} \\ \hline \hline
ROG & 0.7375 & 0.1222 & 0.3581 & 0.7375 & 0.5791
\end{tabular}
\label{tab:attacks}
\end{table}

\subsubsection{Adversarial robustness.}
\begin{figure*}[!t]
\begin{center}
\includegraphics[width=0.95\linewidth]{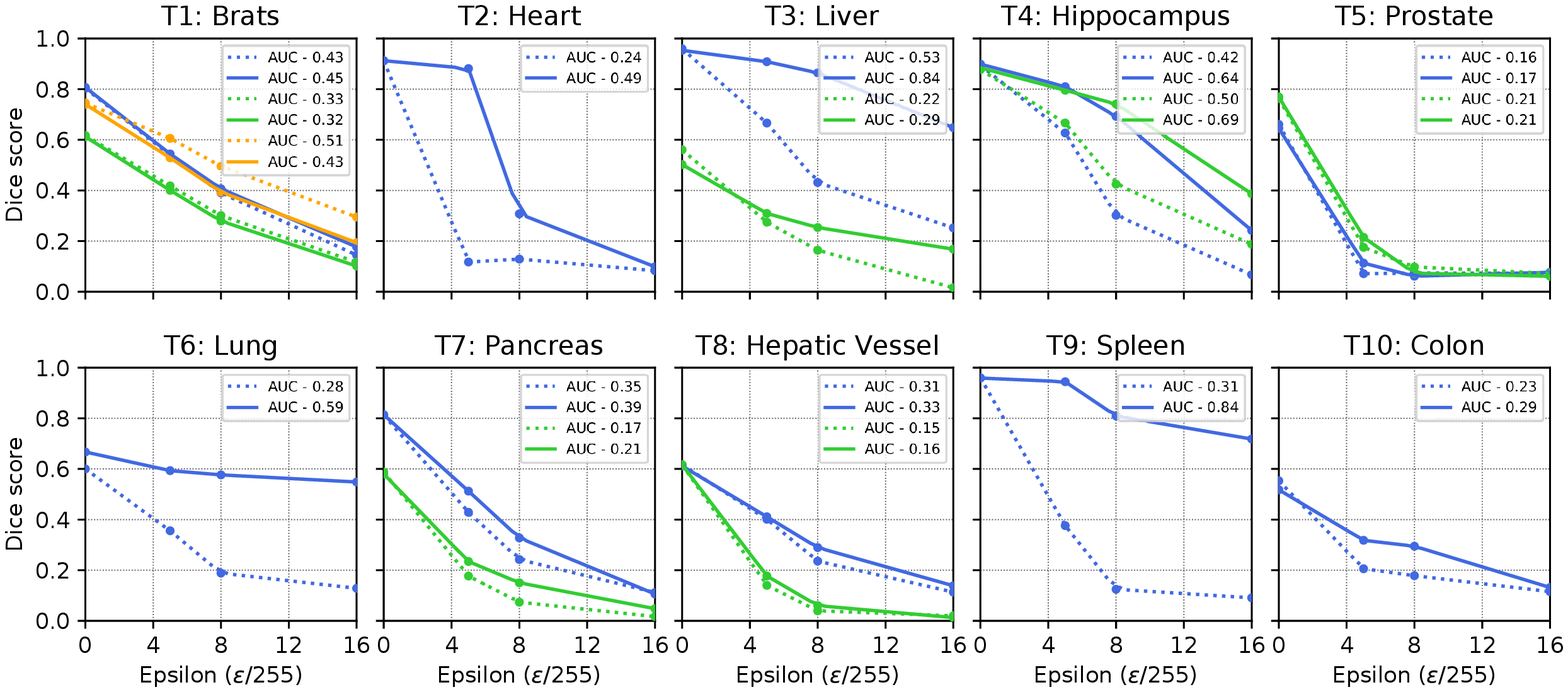}
\end{center}
   \caption{\textbf{Adversarial robustness for the 10 tasks in the MSD.} We train ROG on clean images (dotted lines) and with Free AT (solid lines), and evaluate their adversarial robustness across various $\epsilon$. Every color represents a different category within each task. In general, training with Free AT improves adversarial robustness against attacks.}
\label{fig:tasks}
\end{figure*}

We conduct attacks on every task of the MSD, vary the attack strength ($\epsilon$), and report the Dice score in Figure~\ref{fig:tasks}. Dotted lines represent the performance of ROG with regular training, solid lines show ROG with Free AT, and categories within each task are color-coded. The effect on the Dice score with regular training demonstrates the susceptibility of segmentation models to adversarial attacks. While the ranges for Dice score vary greatly across tasks and categories, our results suggest a coherent finding consistent with the robustness literature: training with Free AT is a tractable approach to improving the adversarial robustness of DNNs. In addition, in most tasks, Free AT enhances robustness \textit{without} major costs to performance on \textit{clean} samples. We identify that the robustness improvement is greater for large structures, such as the liver, and for definite structures like the left atrium of the heart.

\subsubsection{Qualitative evaluation.}

\begin{figure*}[t]
\begin{center}
\includegraphics[width=0.9\linewidth]{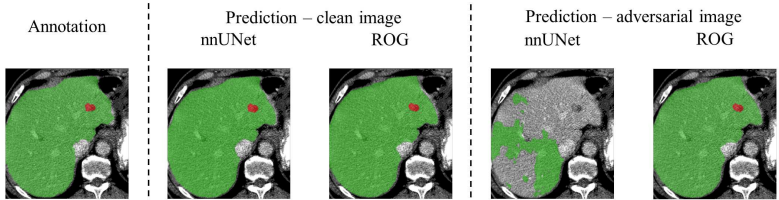}
\end{center}
    \caption{\textbf{Qualitative results.} We compare the results of ROG and nnUNet on liver images. The adversarial examples are obtained using APGD-CE with $\epsilon=8/255$ and $m=5$. Both methods have comparable performance on clean images, while ROG is vastly superior in the presence of adversarial noise.}
    \label{fig:qualitative}
\end{figure*}

We show qualitative results in Figure~\ref{fig:qualitative}, where the task is to segment liver and liver tumors. We compare ROG with nnUNet over clean and adversarial images computed through APGD-CE with $\epsilon = \nicefrac{8}{255}$ and $m=5$. For more results, please refer to the supplemental material. Our results show that, on clean images, ROG provides similar segmentations to those of nnUNet. However, on adversarial images, both methods strongly diverge in their performance. In particular, ROG preserves the segmentations rather untouched, while nnUNet's segmentations are shattered, and the tumor completely disappears. We highlight a visually appealing phenomenon: nnUNet's segmentations appear to follow a pattern of segmenting elements whose boundaries usually correspond with high contrast boundaries. These patterns suggest that nnUNet, and perhaps other segmentation methods, are subject to strong inductive bias that may come from the training data or the convolutional layers with small kernels.

All in all, we observe that nnUNet's vulnerability to adversarial noise is high, failing both on the side of false positives and false negatives. In real-life clinical scenarios, the consequences of such mistakes, while largely unpredictable, could be appalling. Since ROG suffers from these effects to a minor degree, it appears as a better alternative for reliable medical image segmentation.

\subsection{Results on the Medical Segmentation Decathlon}

\begin{table*}[t]
\centering
\caption{\textbf{MSD test set results.} Comparison of Dice score with top performing methods of the MSD. The results were retrieved from the MSD official leaderboard.}
\begin{tabular}{c|c|c|c|c|c|c|c|c|c|c||c|c|}
\cline{2-13}
 & T1 & T2 & T3 & T4 & T5 & T6 & T7 & T8 & T9 & T10 & Mean & Std \\ \hline
\multicolumn{1}{|c|}{MPUNet~\cite{oneNetworkDec}} & 0.6 & 0.89 & 0.75 & 0.89 & 0.77 & 0.59 & 0.48 & 0.48 & 0.95 & 0.28 & 0.66 & 0.22 \\ \hline 
\multicolumn{1}{|c|}{nnUNet~\cite{nnunet}} & 0.61 & 0.93 & 0.84 & 0.89 & 0.83 & 0.69 & 0.66 & 0.66 & 0.96 & 0.56 & 0.75 & 0.15  \\ \hline 
\multicolumn{1}{|c|}{C2FNAS~\cite{c2fnas}} & 0.62 & 0.92 & 0.84 & 0.88 & 0.82 & 0.70 & 0.67 & 0.67 & 0.96 & 0.59 & 0.75 & 0.14 \\ \hline \hline 
\multicolumn{1}{|c|}{ROG (Ours)} & 0.56 & 0.91 & 0.75 & 0.89 & 0.79 & 0.36 & 0.66 & 0.62 & 0.88 & 0.46 & 0.69 & 0.18 \\ \hline 
\end{tabular} 
\label{tab:resultsDec}
\end{table*}

To verify the generalization capability of ROG in a clean setup, we evaluate our results in the test server of the MSD. Table \ref{tab:resultsDec} summarizes our results compared against all published methods that are evaluated over the entire MSD. ROG surpasses the performance of MPUNet in most tasks with nearly $24\times$ fewer parameters ($2.6M$ vs. $62M$). When we compare to more complex models, we see a trade-off between performance and efficiency. While nnUNet uses an ensemble of models, ROG relies on a \textit{single} architecture. In contrast, C2FNAS uses a single fixed model for most tasks, but requires $6.5\times$ more parameters and almost $5\times$ more GFLOPs ($150.78$ \textit{vs.} $30.55$). Note that, since ROG adapts to each dataset, we compute the total number of FLOPs by averaging the results of all methods using a maximum input size of $96 \times 96 \times 96$, as in~\cite{c2fnas}. Under this scheme, our most computationally expensive method has only $79.32$ GFLOPs. Considering the large computational demands of studying adversarial robustness, specially in the domain of 3D segmentation, these results demonstrate the advantage of adopting ROG as a strong baseline for the analysis of this dimension in the task of generic medical segmentation.

\section{Conclusion}
We introduce a benchmark for evaluating the adversarial robustness of volumetric segmentation systems in the medical domain. In particular, we extend the state-of-the-art AutoAttack approach from the image-recognition domain to the volumetric semantic segmentation domain. Further, we propose ROG, a robust and efficient segmentation method for the MSD. We find that ROG provides competitive results over clean images in most tasks of the MSD, while vastly outperforming one of the top-performing methods in this benchmark under adversarial attacks. Our study provides further evidence that the adversarial robustness dimension has been profoundly under-explored in medical image segmentation. Hence, we hope that our work can serve as a tool for developing accurate and robust segmentation models in the clinical domain.

\textit{Acknowledgements:} we thank Amazon Web Services (AWS) for a computational research grant used for the development of this project.
%
%
%
\bibliographystyle{splncs04}
\bibliography{main}
\end{document}